\documentclass{article}%
\usepackage{authblk}
\usepackage[utf8]{inputenc}
\usepackage{amssymb}
\usepackage{amsmath}
\usepackage[T1]{fontenc}
\usepackage{amsfonts}
\usepackage[super]{cite}
\usepackage{graphicx}
\usepackage{subfigure}
\usepackage{algorithm}
\usepackage{algorithmic}
\usepackage[utf8]{inputenc}
\usepackage{textcomp}
\usepackage{eurosym}
\usepackage{color}
\usepackage{url}

\title{A Novel Navigation System for an Autonomous Mobile Robot in an Uncertain Environment}
\author[1]{Meng-Yuan Chen}
\author[2]{Yong-Jian Wu}
\author[3]{Hongmei He \thanks{Corresponding author. E-mail: h.he@cranfield.ac.uk}}
\affil[1]{College of Electrical Engineering, Anhui Polytechnic University, China.\\}
\affil[2]{Wuhu HIT Robot Technology Research Institute Co. Ltd, China.\\}
\affil[3]{School of Aerospace, Transport and Manufacturing, Cranfield University, Cranfield, MK43 0AL, UK.\\}
\date{June 2020}
\begin{document}
  \maketitle

\begin{abstract}
In this paper, we developed a new navigation system, which detects obstacles in a sliding window with an adaptive threshold clustering algorithm, classifies the detected obstacles with a decision tree, heuristically predicts potential collision and finds optimal path with a simplified Mophin algorithm. This system has the merits of optimal free-collision path, small memory size and less computing complexity, compared with the state of the arts in robot navigation. The experiments on simulation and a robot for eight scenarios demonstrate that the robot can effectively and efficiently avoid potential collisions with any static or dynamic obstacles in its surrounding environment.

\end{abstract}

Researchers have thoroughly investigated the mobile robot navigation problems, and achieved a certain progress in this area. As the navigation of autonomous mobile robots with free collisions could increase the range of their applications, thus, it has played important role in various application domains, such as transportation, rescue services, detection, mining, space exploration and military, etc. For example, autonomous vehicles are constantly improving with the autonomous navigation capability \cite{Kocic2018}, and have obtained the qualification on roads in some countries (e.g. Google autonomous vehicle). In the fields of coal mines, intelligent mobile robots have been used to replace the manual work and tedious repetitive operations \cite{Ray2016}. Robots are key to future space exploration \cite{Britt2019}, and the autonomous navigation of mobile robots allows them to have better operation capabilities. In the military field, mobile robots with autonomous navigation can better help reach military missions, such as military patrols on the desert border \cite{Joshi2019}. For the success of all applications above, effectively autonomous navigation of robots is the precondition and guarantee.

Autonomous navigation requires a robot to be able to precept the surrounding environments through processing or fusing the data, collected from sensors, and highly performed robot perception enables a robot to make a right decision and thus to have a right response to any anomaly situation in its surrounding environment.
Autonomous path planning with obstacles avoidance in dynamic environments is a crucial issue in the navigation of a robot \cite{WangJ2015,LuoJ2017,ZhangW2017}. There are some classical obstacle avoidance methods, such as the Artificial Potential Field (APF)\cite{Khatib1985}, the Virtual Force Field (VFF) \cite{Borenstein1989} and the Vector Field Histogram (VFH) \cite{Borenstein1991}. APF provides a simple and effective motion planning method. The basic idea is to treat the robot's configuration as a point in a potential field that combines attraction to the goal and repulsion from obstacles, hence, the resulting trajectory is output as the path. However, APF has a major problem that the robot is easy to be trapped at a local minimum before reaching its goal \cite{Weerakoon2015}.  Borenstein and Koren \cite{Borenstein1989} proposed the VFF method, which uses a two-dimension Cartesian histogram grid for obstacle representation. Each cell in the histogram grid holds a certainty value, which represents the confidence of the algorithm in the existence of an obstacle at that location. As VFF could produce oscillatory and unstable motion in some cases, Borenstein and Koren \cite{Borenstein1991} further proposed the VFH method, which uses a two-dimensional Cartesian histogram grid as a world model. This world model is updated continuously with range data sampled by on-board range sensors. In the first stage of VFH, the histogram grid is reduced to a one-dimensional polar histogram that is constructed around the robot's momentary location. Each sector in the polar histogram contains a value representing the polar obstacle density in that direction. In the second stage, VFH selects the most suitable sector from all polar histogram sectors with a low polar obstacle density, and the steering of the robot is aligned with that direction. A threshold adapting to the relationship between the obstacle's position and the target point was used in the improved vector field histogram avoidance algorithm \cite{LiuJ2015}. However, the obstacles in all the studies are limited to static obstacles, regardless of dynamic obstacles, and these methods did not have a further stage to optimise the path for obstacle avoidance.

 We have proposed a complete autonomous navigation system of a robot, named as ATCM \cite{ChenMY2018}. In this research, we further improve the new navigation system with the six stages from data collection, obstacle detection, obstacle classification, potential collision prediction, optimal path planing, to the robot's behaviour updating. A clustering algorithm with an adaptive threshold \cite{ZhongX2011} is developed for detecting various shapes of obstacles, a simple decision tree is used to classify the detected obstacles in terms of the coordinates and dynamics of the detected obstacles, a heuristic algorithm is provided for potential collision prediction in terms of the dynamics of the robot and obstacles, and finally, the Morphin algorithm \cite{Simmons1996} is simplified to find an optimal path without collision.

One of main objectives for robot navigation is to improve the navigation performance. While a robot can avoid any static or dynamic obstacles on its paths, the following three performance indicators are often used to evaluate the navigation of a robot:

{\it Positioning accuracy}, including static single point positioning error and dynamic track error, where single point positioning error consists of $X$ error, $Y$ error and total error maximum, expectation, entropy and super entropy, and dynamic trajectory error includes robot direction error, robot position maximum error, error mean, error expectation, error entropy and error hyper-entropy. He et al. \cite{He2014} used linguistic decision tree (LDT) for robot routing problem, and the fuzzy technology in the LDT enables the robot to effectively overcome the stochastic errors due to such factors as mechanical properties of robot's motor and the friction changing on the floor even in a static environment, etc, thus reducing the accumulated error of robot's position.

{\it Speed index}, including the maximum speed  or average speed of the mobile robot during the whole test process. Especially the maximum speed depends on the computing complexity and mechanical properties of a robot. Under the mechanical properties of a robot, the computing complexity of robot navigation determines the maximum speed, representing the real-time performance.

{\it Navigation efficiency}, including the length of time $T$ used by the mobile robot to complete a specific task, which is related to the speed $V$, the total length $L$ of the entire process trajectory. Under a specific speed $V$, the length of the path that robot travels from a start point to the target represent the navigation efficiency.

The navigation of a robot should be robust in respects of the performance indicators above, reflecting on all steps of the whole process. However, most of existing research addressed partial steps of a navigation process. There was little investigation of the whole navigation system. This is not helpful for evaluating the performance of the whole process of a robot’s navigation.

Hence, in this paper, we examine how all the steps work harmonically and time changes for 8 scenarios, which are typical scenarios in real dynamic world.  The navigation system is simulated and integrated in a physical robot. The experiments are conducted on the simulation system and a physical robot.

This paper is organised as follows: Section 2 reviews the state of the arts in robots' navigation; Section 3 provides the details of the proposed navigation system; Section 4 provides three simulation experiments, time performance test for 8 real scenarios, and the validation on a physical robot in a dynamic environment with multiple static and obstacles; finally, some conclusions and future work are given in Section 5.
\section{Existing work}
Navigation of mobile robots is a classic issue in robotics. The robot's route learning problem is one of robot's navigation problems. Conventionally, an NARMAX model, a non-linear system identification approach, was trained to represent the sensor-motor task \cite{Gardiner2012}. To improve the robustness of the robot's navigation and overcome the drawback of NARMAX in losing performance due to the dynamics of a running robot, a linguistic decision tree (LDT) was developed for the robot routing problem \cite{He2014}, and achieved excellent performance. In fact, both of the two methods are guided by human, in which, the data are retrieved when human drives the robot along a specific path in a fixed environment, and then the model (e.g. NARMAX or LDT) is trained. Hence, this work is suitable for a robot working on a specific task in a fixed environment.

However, for most cases, we expect robots can work in an uncertain environment with free collision. There has been much research in this area. Generally, it can be categorised to map-based navigation and mapless navigation. Whereas map-based navigation can be subdivided in metric map-based navigation and topological map-based navigation, mapless navigation can include reactive techniques based on qualitative characteristics extraction, appearance-based localization, optical flow, features tracking, plane ground detection/tracking, etc. \cite{Bonin2008}. A grid-based mapping technique is usually used in the map-based navigation system for an autonomous mobile robot \cite{Cho1995}.

SLAM (Simultaneous localization and mapping) is a technique, used by robots and autonomous vehicles to build a map within an unknown environment, or to update a map within a known environment, while keeping track of their current locations \cite{Durrant-Whyte2006,Bailey2006}. Various SLAM techniques have been developed. For example, Kovacs et al. \cite{Kovacs2019} proposed a landmark-selection approach based on explicit and spatial information for solving mobile robot navigation problems, where, the visual odometry-based motion estimation supports the template matching of landmarks. The important characteristic of SLAM techniques that could assist in autonomous navigation is the ability of a mobile robot to concurrently construct a map for an unknown environment and localize itself within the same environment. However, a SLAM system might fail in handling extremely dynamic or harsh environments. Storing the map during long-term operation is still an open problem. Even when the data is stored on the cloud, raw data points or volumetric maps may cost much memory; similarly, storing feature descriptors for vision-based SLAM quickly becomes burdensome \cite{Cadena2016}.

To implement a robot's autonomy, its perception to external environments is very important. Advanced sensors provide enabling techniques for robots' perception. Most frequently used sensors include laser sensors \cite{ZhangY2016,ZhangD2016,He2014}, visual sensors \cite{ZhangQ2013}, infrared sensors \cite{WangM2016} and ultrasonic sensors \cite{WangC2015}. Different sensor techniques may decide different navigation techniques.

Now, laser sensors are increasingly used, as they have the advantages of wide  detection range, high-precision measures, high reliability, good stability, strong anti-interference and light weight.  Recently, Xin et al. \cite{XinY2014} developed a dynamic obstacle detection model by fusing  the Velodyne data from a 3D laser sensor and the motion state information from a 4-wire laser sensor to derive the position of a moving obstacle in a grid map based on the confidence distance theory.  In \cite{YangY2013}, an approach to detecting the speed and direction of an obstacle was proposed, and the obstacle avoidance was implemented, regarding the least Euclidean distance from the robot to the edges of the obstacle. As only circular objects were addressed. This method is not robust for the diversity of obstacles.

Vision techniques have played significant roles in robots' navigation \cite{Bonin2008}. For example, in the early stage, the highly notable planetary vehicle, Mars Pathfinder \cite{Matthies1995}, used a stereo camera to shoot images of the Mars surface, when the rover explores the environment, which is controlled by human through selecting the goal point in 3D representations of previously captured images of the terrain.  Further, pattern recognition from images have been widely used for obstacle detection in autonomous robot navigation \cite{Chen2000,Viet2007}.

Artificial Intelligence techniques are important drivers for implementing the autonomy of a robot. Especially, machine learning techniques have been widely applied for obstacle detection. For example, neural networks have been developed for a robot's path planning \cite{Jung1999,Ouarda2011}. A support vector machine based on the space-time feature vector was developed to recognize dynamic obstacles, but without the further exploration of the path planning for obstacle avoidance \cite{HuangR2016}. Wang et al. \cite{Wang2018} proposed a geomagnetic gradient bionic model with a parallel approach for robot navigation, which becomes a multi-objective convergence problem. Mohanta and Keshari \cite{Mohanta2019} proposed a knowledge-based fuzzy control system for target search behavior and path planning of mobile robots. This approach includes two stages: the first stage is to generate a shortest path between the starting position and the target position in a previously known messy environment, wherein the probability road-map is used to construct a straight path by connecting the intermediate nodes; the second folding step helps to convert the sharp angle to a smooth curve throughout the path.

It is commonly believed that deep convolutional neural networks (CNNs) are good for vision-based pattern recognition problems. Hence, CNNs have been applied to improve the perception of robots. For example, Steccanella et al. \cite{Steccanella2020} proposed a two-stages approach to detecting  waterline and obstacles under the detected waterline based on images from low-cost autonomous boats, with CNN and vision techniques respectively,  for environmental monitoring, and their statement of the computing at 10 frames per second on an embedded GPU board indicates the computing cost of the proposed vision-based method; Wu et al. \cite{WuP2017} proposed a two-stages path planning method based on a CNN. Firstly, the comprehensive features are extracted directly from original images of roads through a CNN; then, robots determine their moving direction in terms of the classification results from the CNN; This approach is good for a static environment, but not good for dynamic environments. Hence, it could be applied for robots' pre-training to get an initial path in an static environment. Due to the computing complexity, the first stage is completed off-line. Online neural network training to adapt the dynamic environment is still an open question.

Various meta-heuristic algorithms have been developed for robot path planning as well. Usually, these approaches transform the path planning problem to an optimisation problem to implement offline planning by producing a shortest path in an static environment with stationary obstacles, such as the ant colony algorithm \cite{Brand2010,Rashid2016}, Genetic Algorithms \cite{Jiang2014}, the simulated annealing based approach \cite{Ganeshmurthy2015} and Particle Swarm Optimisation (PSO) \cite{Ahmed2015}. As they are time consuming, off-line planning is needed. Based on the path generated by off-line planning, the robot travels through the stationary obstacles, then the robot will recalculate the path when a dynamic obstacle appears on the path. The issue of such two stage approaches is that the remaining path might not be the best after recalculating the path to pass the detected dynamic obstacle, especially in dense obstacle environments.

Herojit Singh and  Khelchandra \cite{Herojit2019} proposed a mobile robot navigation approach based on fuzzy genetic algorithm (GA). Information about the distance and angle of obstacles from the robot is obtained by the exploration of three directions in front of the robot ($40^\circ$, $0^\circ$ and $-40^\circ$). When all three paths are blocked by obstacles, the fuzzy system is used to avoid obstacles; otherwise, the conflict-free path is selected from the three directions. The GA is used to find the optimal range of linguistic values of the variables of the membership function. It is shown that fuzzy-GA with the three-paths concept is computationally efficient, compared to other hybrid methods. Wan et al. \cite{WanX2015} proposed an improved Ant Colony Algorithm, combining the Morphin algorithm for mobile robot path planning. In the process of the global optimisation, the Morphin algorithm is applied for local path planning in each step.

The Morphin algorithm was proposed based on the Ranger algorithm by Simmons \cite{Simmons1996} for obstacle avoidance and safeguarding of a lunar rover in space. The Morphin path planning is an area-based approach. The basic idea is that the patches of terrain are analysed to determine the traversability of each patch. The Morphin algorithm has the merits of low computing complexity and thus it can implement the real-time performance for a robot's path planning \cite{Zhang2019}. Hence, the Morphin algorithm has been widely applied for path planning in an uncertain environment with dynamic obstacles\cite{ZhuGe2014,WanX2015}, and it, integrating in a global path planning algorithm, has achieved good performance for the path planning in indoor and complex environments \cite{Wu2018,Zhang2016}.

Various approaches for robot path planning have been developed by researchers \cite{Kamil2015}, and obtained significant performance in solving certain aspects of the path planning problem. However, they have their own disadvantages \cite{Tang2012,Motlagh2012}. Therefore, a high-performance autonomous navigation system in effectiveness and efficiency is still demanded.
\section{The Proposed System}
\begin{figure}
\centering
\includegraphics[height=2.0in]{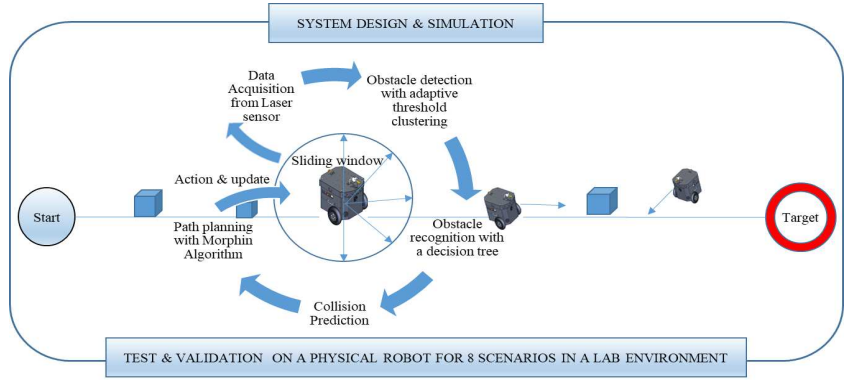}
\caption{The flow chart of the navigation system} \label{fig:navigationsetps}
\end{figure}

A navigation system of a mobile robot is thoroughly investigated, regarding all the six steps shown in Fig. \ref{fig:navigationsetps}.
Initially, the robot sets a sliding window, the center of which is the robot; the environmental data is collected from the laser scanner on the robot; static and/or dynamic obstacles in the surrounding environment are detected based on the data within the window, using the adaptive-threshold clustering algorithm; a simple decision tree, learned from experienced parameters, is used to determine the types of obstacles (new, dynamic or static); the movement of dynamic obstacles is computed, in terms of their speed and direction and the relation between the robot and the detected obstacles, and the potential of conflict is predicted; the Morphin algorithm is applied to avoid obstacles if a potential collision in front of the robot is not avoidable; finally, the robot updates its state, moving toward the generated local sub-target, and correspondingly, the sliding window is updated; the process is repeated until the robot reaches the global target. Algorithm \ref{alg:navigation} provides the pseudo-code of the navigation System.
\begin{algorithm}[ht]
\caption{RobotNavigation($r_w$, $R$, $Target$)}\label{alg:navigation}
\begin{algorithmic}
\STATE Initialise($r_w$, $R$); /*a sliding window with radius of $r_w$ */\\
\STATE t=0; \\
\WHILE {($R$ has not reached the target)}
    \STATE $D$=ReadData(); /*from laser scanner*/\\
    \STATE $O_{chain}$ = Clustering ($D$); \\
    \FOR  {($O_k(t)\in O_{chain}(t))$}
        \STATE $O_{type}$ =ObstacleRecognision($O_k{t}, O_{chain}(t-1)$);\\
        \STATE $O_{info}$=calculateObstacleInfo($O_k(t)$,$O_{type}$);\\
        \STATE $Collision$ = CollisionPrediction($O_{info}$, $R$); \\
        \IF {($Collision \neq NULL $)}
            \IF{($Collision$ is avoidable)}
                \STATE $V_R=V_R-\Delta V_R$; /*slow done the robot*/
            \ELSE
                \STATE Mophin($Collision$, $R$); \\
            \ENDIF
            \STATE break;
       \ENDIF
    \ENDFOR\\
    \STATE update($R$, $r_w$);
    \STATE $t = t+1$; \\
\ENDWHILE
\end{algorithmic}
\end{algorithm}
\subsection{Data acquisition from the laser sensor}
A Grid Map with a specified resolution is usually created in map-based navigation \cite{ZhuJ2015}. In this research, a grid based environment model is established. Assume the robot is in the global coordinate system, sharing the origin of the global coordinate system. To collect the information of obstacles in the surround environment of the robot, a two-dimensional laser sensor, the product of the German company SICK, is installed on the robot. The data from the laser scanner on the robot is proportion to the distance between a robot to an obstacle, which reflects the time interval between sending and return of the laser beams. In each step of data collection, the laser scanner scans the front semicircle of the robot, ranging in [0$^{\circ}$, 180$^{\circ}$], with a angular interval of $0.5^{\circ}$.

\begin{figure}
\centering
\includegraphics[height=2in]{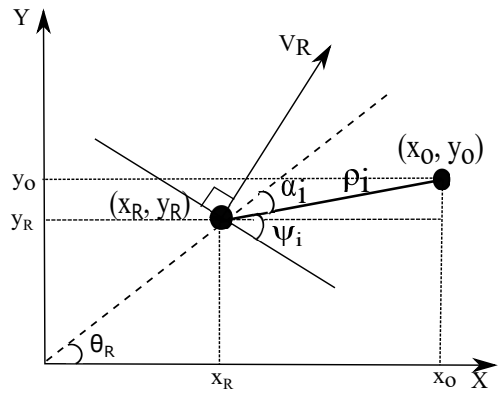}
\caption{The Robot Model} \label{fig:RobotModel}
\end{figure}

Fig. \ref{fig:RobotModel} illustrates the positions of a robot and an obstacle and their relationship in positions. The pair of ($\rho_i$, $\alpha_i$) represents the location of an obstacle in the local polar coordinates, where the robot is the original point, $\rho_i$ indicates the length of a laser beam, representing the distance between the obstacle and the robot, $\varphi_i \in$ [0$^{\circ}$,180$^{\circ}$], $i$ = 0...360, indicating the index of the laser beam; $\alpha_i$ is the angle between a laser beam and the robot's direction, $\theta_R$ is the angle of the robot in the global polar coordinates. The coordinates of the obstacle's position in the global coordinate system can be calculated with Eq. (\ref{eq:globallocation}). The global coordinates ($x_o, y_o$) of an object can be transferred to the coordinates ($x'_o, y'_o$) in the grid map with Eq. (\ref{eq:gridlocation}).
\begin{align}\label{eq:globallocation}
x_o=x_R+\rho_i cos(\theta_R-\alpha_i), \nonumber \\
y_o=y_R+\rho_i sin(\theta_R-\alpha_i).
\end{align}
\begin{align}\label{eq:gridlocation}
x'_o = floor(\frac{x_o}{r}+\frac{1}{2}), \nonumber\\
y'_o = floor(\frac{y_o}{r}+\frac{1}{2}),
\end{align}
where, $r$ is the resolution of the grid map (see Fig. \ref{fig:GridMap}).
\subsection{Data Clustering for Obstacle Detection}
\begin{figure}
\centering
\includegraphics[height=2in]{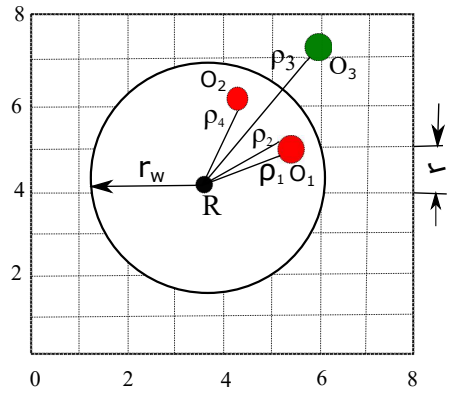}
\caption{The Grid Map Model} \label{fig:GridMap}
\end{figure}

After collecting the data from the laser scanner, the robot will check if there exist any obstacles within the sliding window. An adaptive threshold nearest neighbor clustering method is developed to group data points. It is believed that the data out of the sliding window (i.e. $\rho > r_w$) does not provide any hazard to the robot. Hence, it will not be used for clustering, like $O_3$ in Fig. \ref{fig:GridMap}, as $\rho_3>r_w$. The Euclidean distance is used to represent the distance between two data points. Two available consecutive data points ($\rho_2$ and $\rho_4$) are believed to belong to different obstacles, respectively, if the distance between them is larger than the distance between two neighbouring laser beams that have the identical length and the angular interval 0.5 (e.g. $\rho_1$ and $\rho_2$).
Therefore, a threshold $\theta$ is defined, linearly related to the value of [$\rho(t) sin (0.5)$], which is the approximate of the distance between two neighbouring data points in one cluster. $\theta$ is proportional to the current laser beam $\rho(t)$.  As the shape of an obstacle could be irregular, we use an adaptive rate $\lambda$ to indicate the irregularity of obstacle shapes. Due to the introduction of $\lambda$, two closed obstacles could be viewed as one cluster. This could improve the robustness of the clustering algorithm. For the rectangle of obstacles, $\lambda$ can be set to close to 1.
\begin{align}\label{eq:theta}
\theta = \lambda\rho(t)sin(0.5^{\circ});
\end{align}
To cluster the data points, one can calculate the distance between the current data point and the previous data point in turns of the data points from $0^\circ$ to $180^\circ$. If the distance is larger than the threshold, then the current data point belongs to a new cluster, otherwise, it belongs to the cluster that the previous data point belongs to.  A cluster represents an obstacle. Finally, the clustering algorithm will produce a chain of obstacles $O_{chain}$ (Eq. (\ref{eq:obstaclechain}) at time $t$.
\begin{align}\label{eq:obstaclechain}
O_{chain}=\{O_1(t),O_2(t),... ,O_n(t)\}.
\end{align}
The attributes of an obstacle, $O_k(t)$), can be represented as the vector of four items in Eq. (\ref{eq:obstacleinfo}).
\begin{align}\label{eq:obstacleinfo}
O_k(t)=(Z_k (t),S_k (t),\xi_k(t),V_k(t)),
\end{align}
where, $Z_k(t)$ is the center of $O_k(t)$, $S_k(t)$ is the area of the grids that $O_k(t)$ occupies, $\xi_k(t)$ is the coincidence of $O_k(t)$ to an obstacle at time $t-1$, and $V_k(t)$ is the speed of $O_k(t)$. If $O_k(t)$ is a dynamic obstacle, then $V_k(t)$>0, otherwise, $V_k(t)$=0. Assume $O_k(t)$ is represented by cluster $C_k$, which includes $n_k$ laser beams, {$l_1$, …, $l_{nk}$}, and $l_i$ = ($\rho_i,\alpha_i$).
The center $Z_k(t)$ of $O_k(t)$ can be calculated with Eq. (\ref{eq:laserbeam}), and the global coordinates of the center can be calculated with Eq. (\ref{eq:globallocation}).
\begin{align}\label{eq:laserbeam}
&\overline{\alpha}_k =\frac{\sum_{i=1}^{n_k}\alpha_i}{n_k}, \nonumber\\
&\overline{\rho}_k  = \frac{\sum_{i=1}^{n_k}\rho_i}{n_k}.
\end{align}
It is easy to obtain the pairs ($min(\rho)$, $min(\alpha)$) and ($max(\rho)$, $max(\alpha)$)) in cluster $C_k$, which determine the edges of cluster $C_k$. The global coordinates of cluster edges  ($x_{k,min}$, $y_{k,min}$), and ($x_{k,max}$, $y_{k,max}$) can be calculated with Eq. (\ref{eq:globallocation}); Further, the grid coordinates of cluster edges,($x'_{k,min}$, $y'_{k,min}$), and ($x'_{k,max}$, $y'_{k,max}$) can be calculated with Eq. (\ref{eq:gridlocation}). Hence, the area $S_k$ covers all grids within the coordinates ranges, expressed by Eq. (\ref{eq:range}).
\begin{align}\label{eq:range}
&x'_k \in [x'_{(k,min)} ,x'_{(k,max)}],\nonumber\\
&y'_k \in [y'_{(k,min)} ,y'_{(k,max)}].
\end{align}
One can calculate the global and grid coordinates of all data points in $C_k$ at time $t$ with Eq. (\ref{eq:globallocation}) and (\ref{eq:gridlocation}), respectively.
Assume obstacles $O(t)$ and $O(t-1)$ occupy areas $S(t)$ and $S(t-1)$, respectively. The grid coordinates of $S(t)$ and $S(t-1)$ are shown in Eq. (\ref{eq:t}).
\begin{align}\label{eq:t}
&x(t)   \in [x_{min}(t) ,x_{max}(t)],\nonumber\\
&y(t)   \in [y_{min}(t) ,y_{max}(t)],\nonumber\\
&x(t-1) \in [x_{min}(t-1) ,x_{max}(t-1)],\nonumber\\
&y(t-1) \in [y_{min}(t-1) ,y_{max}(t-1)].
\end{align}
If $S(t)$ and $S(t-1)$ fully overlap, the ($x_{min}, y_{min}$) and ($x_{max}, y_{max}$)  of the two areas should be the same. Hence, Eq. (\ref{eq:fulloverlap}) is true.
\begin{align}\label{eq:fulloverlap}
&S(t)\cap S(t-1) = S(t), \nonumber\\
&S(t)\cup S(t-1) = S(t).
\end{align}
If $x_{max}(t-1) < x_{min}(t)$ or $x_{min}(t-1) > x_{max}(t)$ or $y_{max}(t-1) < y_{min}(t)$ or $y_{min}(t-1) > y_{max}(t)$, then the two areas do not overlap, otherwise, the two areas overlap partially or fully.

If $S(t)$ and $S(t-1)$ partially overlap, their $x$ ranges and $y$ ranges overlap. Hence, one can sort the boundary values of $x$ at times $t$ and $t-1$, and the boundary values of $y$ at times $t$ and $t-1$. Then the order of $x$ boundary values $x_{b1}\leq x_{b2}\leq x_{b3}\leq x_{b4}$ and the order of $y$ boundary values $y_{b1}\leq y_{b2}\leq y_{b3}\leq y_{b4}$ can be obtained. Hence, the overlapping area has the boundary of $[(x_{b2}, x_{b3}), (y_{b2}, y_{b3})]$, and the number of overlapping grids  can be calculated with Eq. (\ref{eq:overlap-grids}).
\begin{align}\label{eq:overlap-grids}
 S(t)\cap S(t-1)= (x_{b3}-x_{b2}+1)(y_{b3}-y_{b2}+1);
\end{align}
The total number of grids that the two obstacles occupy can be calculated with Eq. (\ref{eq:sum-twoareas}).
\begin{align}\label{eq:sum-twoareas}
 S(t)\cup S(t-1)= S(t)+S(t-1)-S(t)\cap S(t-1).
\end{align}


Further, the coincidence of $O_k$ can be expressed as Eq. (\ref{eq:ObstacleCoincidence}).
\begin{align}\label{eq:ObstacleCoincidence}
\xi_k (t)=\frac{S_{(t)}\cap S_{(t-1)}}{S_{(t)}}.
\end{align}
The spatial correlation ($\varsigma_{k_1,k_2}$) between two obstacles can be expressed as a function of two parameters: the distance ($\delta$) between the centres of two clusters and the non-overlapping rate ($\eta$), and $\varsigma_{k_1,k_2}$ can be calculated with Eq. (\ref{eq:spatialcorrelation}),in which, $\delta$ and $\eta$ are expressed with Eq. (\ref{eq:twoparasbetweenobstacles}), and $y_\delta$ and $y_\eta$ are the efficiencies.
\begin{align}\label{eq:spatialcorrelation}
\varsigma_{k_1,k_2}=&\varsigma(O_{k_1}(t),O_{k_2}(t-1)) \nonumber \\
                   =&y_\delta \frac{1}{\delta+1} + y_\eta  \frac{1}{\eta+1}.
\end{align}
\begin{align}\label{eq:twoparasbetweenobstacles}
\delta&=||Z_{k_1}(t),Z_{k_2}(t-1)||, \nonumber\\
\eta&=1-\frac{(S_{o_{k_1}}(t)\bigcap S_{O_{k_2}}(t-1))}{(S_{o_{k_1}}(t)\bigcup S_{O_{k_2}}(t-1))}.
\end{align}
If $O_k(t)$ and $O_k(t-1)$ represent the same obstacle, the distance between their centres should be zero (i.e. $\delta=0$). If two obstacles fully overlap, $\eta=0$; If two obstacles are isolated, $\eta = 1$.  Assume $y_{\delta}$=0.5 and $y_\eta$=0.5. Two fully overlapped obstacles have $\varsigma$=1. The maximal value in all spatial correlations between $O_k(t)$ and all $O_{k_2}(t-1) \in O_{chain}(t-1)$ is denoted as $\varsigma_{k(t),max}$, and expressed in Eq. (\ref{eq:maxSC}). The maximal spatial correlations of $O_k(t)$ is one of important parameters for distinguishing the type of the obstacle.
\begin{align}\label{eq:maxSC}
\varsigma_{k(t),max}=max_{k_2=1..n_k(t-1)}\varsigma_{k(t),k_2(t-1)}.
\end{align}
\subsection{Recognition of Obstacle Types}
\begin{figure}
\centering
\includegraphics[height=1.6in]{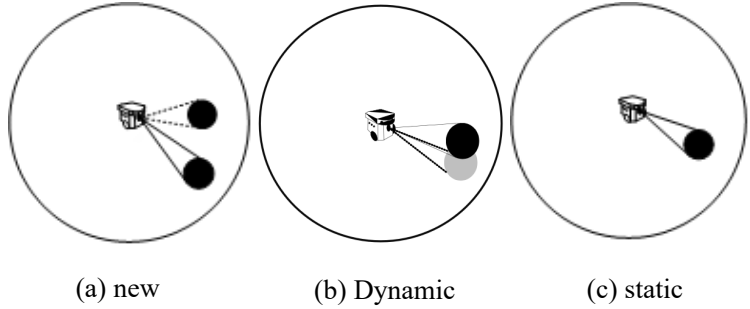}
\caption{The types of obstacles} \label{fig:ObstacleTypes}
\end{figure}

\begin{figure}
\centering
\includegraphics[height=2.4in]{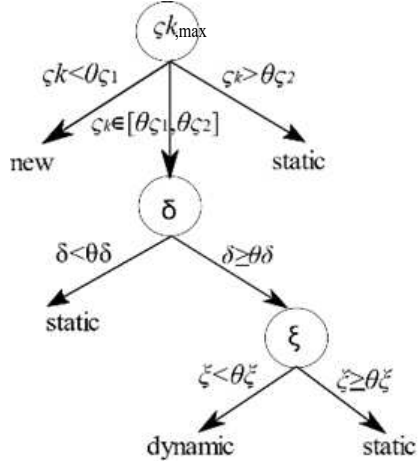}
\caption{Obstacle Recognition with Decision Tree} \label{fig:ObstacleClassification}
\end{figure}

For each obstacle, $O_k(t)\in O_{chain}(t)$, one can easily calculate the center $Z_k(t)$, the grid area $S_k(t)$, and the coincidence $\xi_k(t)$ with Eqs. (\ref{eq:laserbeam})- (\ref{eq:ObstacleCoincidence}), as well as the spatial correlation with Eqs. (\ref{eq:spatialcorrelation}) and (\ref{eq:twoparasbetweenobstacles}). Further, one can identify the maximal spatial correlation $\varsigma_{k(t),max}$ of obstacle $O_k(t)$ to the obstacles in $O_{chain}(t-1)$. Fig. \ref{fig:ObstacleTypes}(a)-(c) show three types of obstacles: new, dynamic and static.
To recognise which type of obstacle is met by the robot in its surrounding environment, a simple decision tree is used, as shown in Fig.\ref{fig:ObstacleClassification}. In the decision tree, the first attribute node is the maximal spacial correlation $\varsigma_{k,max}$ of the obstacle $O_k(t)$. Two thresholds $\theta_{\varsigma_1}$ and $\theta_{\varsigma_2}$ of the spatial correlation are set to recognise whether an obstacle is new or static,  where, (0<$\theta_{\varsigma_1}$ < $\theta_{\varsigma_2}$<1). If $\varsigma_{k,max} < \theta_{\varsigma_1}$, then it is a new obstacle (Fig. \ref{fig:ObstacleTypes} (a));if $\varsigma_{k,max} > \theta_{\varsigma_2}$, then the obstacle is static (Fig. \ref{fig:ObstacleTypes} (c)); if $\varsigma_{k,max} \in [\theta_{\varsigma_1}, \theta_{\varsigma_2}]$, then it might be a dynamic obstacle (Fig. \ref{fig:ObstacleTypes} (b)) or not. So, it will be further assessed in terms of the distance $\delta$ ($0 \leq \delta <r_w$) between two obstacles' centres. A threshold $\theta_\delta$ of the centre distance is set to further judge whether the obstacle is static or not. If $\delta< \theta_\delta$, it can be recognised as a static obstacle,  otherwise, a threshold $\theta_\xi$ of the obstacle coincidence $\xi_{k(t)}$ is used to further judge whether the obstacle is static or dynamic. If $\xi_{k(t)} < \theta_\xi$ , then it is dynamic, otherwise, static. Here, all threshold values are obtained through primary experiments.
\subsection{Movement of a dynamic obstacle}
\begin{figure}
\centering
\includegraphics[height=2.40in]{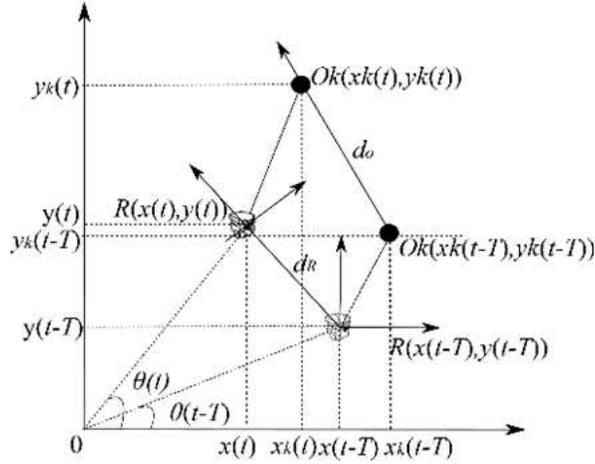}
\caption{Motion process of an obstacle} \label{fig:ObstacleMotion}
\end{figure}

\begin{figure}
\centering
\includegraphics[height=4.4in]{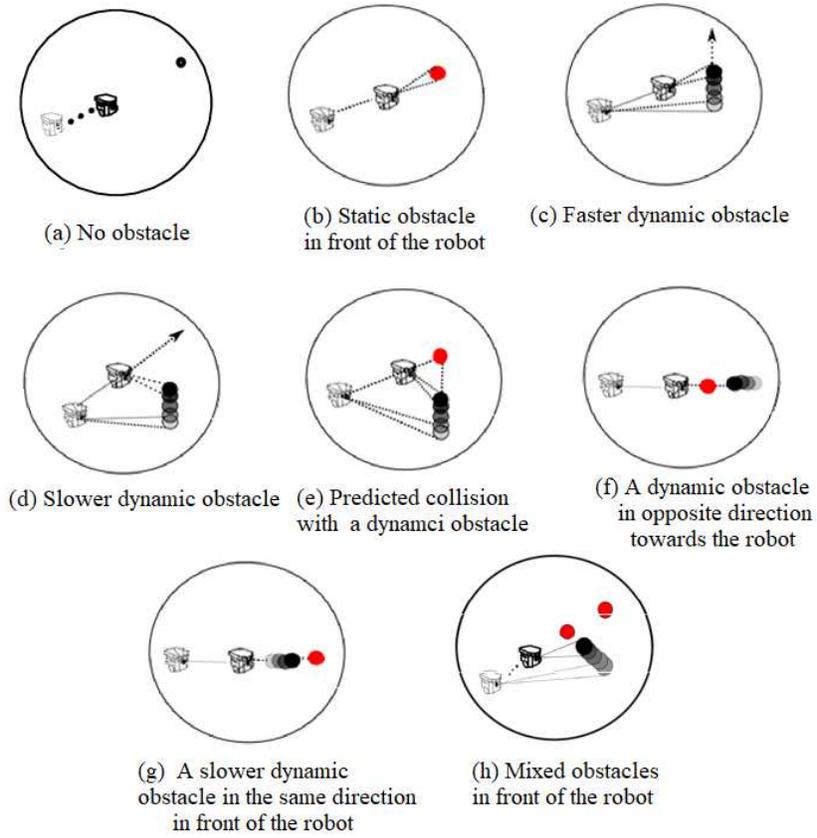}
\caption{Eight scenarios} \label{fig:scenarios}
\end{figure}

If the obstacle is dynamic, the speed and angle of its movement will be further calculated.  Fig.\ref{fig:ObstacleMotion} illustrates the motion process of the obstacle and the robot from $t-T$ to $t$ in the global coordinates system, where, $T$ is the time period of laser scanning from $0^{\circ}$ to $180^{\circ}$.  A moving robot in the environment has the global coordinates $R(x(t), y(t))$ at time $t$ and $R(x(t-T), y(t-T))$ at time $t-T$. One can easily calculate the global coordinates of the obstacle, $O_k(x_k(t), y_k(t))$ at time $t$ and $O_k(x_k(t-T), y_k(t-T))$ at time $t-T$, in terms of the values of laser beams at time $t$ and $t-T$, via Eq.(\ref{eq:globallocation}).
It is easy to calculate the speed $v_o$, the direction angle $\alpha_o$ and the distance $d_o$ of the moving obstacle from $t-T$ to $t$ via Eq. (\ref{eq:ObstacleMovement}). In the same way, one can calculate the speed $v_R$, the direction angle $\alpha_R$ and the distance $d_R$ of the moving robot from time $t-T$ to $t$.
\begin{align}\label{eq:ObstacleMovement}
v_o&= \frac{d_o}{T}, \nonumber \\
\alpha_o&= \arctan (\frac{x_k(t)-x_k(t-T)}{y_k(t)-y_k(t-T)}), \nonumber \\
d_o&=\sqrt{(x_k(t)-x_k(t-T))^2+(y_k(t)-y_k(t-T))^2}.
\end{align}
The potential collisions ahead can be predicted in terms of the states of the robot and the obstacle. There could be eight scenarios where a robot is running, as shown in Fig. \ref{fig:scenarios}. Scenario (a) is the simplest that there is no obstacle, where, the robot is running in the straight direction towards the target; In scenario (b), an obstacle is statically staying on the path where the robot is going, and the potential collision is just at the place where the obstacle is, if the robot does not change its path;  In scenario (c), an obstacle at the probing area of the robot may cross the path of the robot before the robot arrives the crossing point, which may be a potential collision point if robot speeds up; in scenario (d), an obstacle at the probing area of the robot may cross the path of the robot, but it has not arrived at the potential collision point when the robot arrives there; in scenario (e), the robot would collide with the obstacle, when they arrive at the crossing point between the robot's path and the obstacle's path at the same time; in the scenario (f), the obstacle is moving on the path of the robot but towards the robot, hence, the robot would collide with the obstacle, if it does not change its path; in scenario (g), the robot and the obstacle are moving on the same path, and the robot is behind the obstacle, but faster than the obstacle, then the robot would collide with the obstacle at a point on the path if the robot neither deduces its speed nor changes its path; scenario (h) shows both dynamic and static obstacles are on the path of the robot. This requires robot could avoid both obstacles on the path in real-time, no matter how close the two obstacles are.
\subsection{Obstacle collision avoidance}
\begin{figure}
\centering
\includegraphics[height=1.8in]{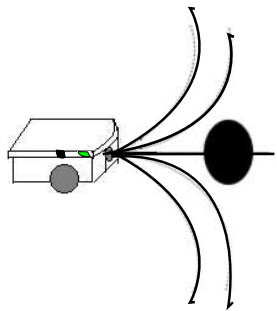}
\caption{Morphin Paths} \label{fig:morphin}
\end{figure}

The classic Morphin algorithm is used to implement the local path planning based on limited environment information. If a robot identifies an obstacle, and finds that it might collide with the obstacle, then it will set up a few alternative paths, and select the best path in the alternative paths to replace the path where the robot and the obstacle will meet, thus avoiding the obstacle (Fig. \ref{fig:morphin}).
In the eight scenarios (Fig. \ref{fig:scenarios}), scenarios (b),(e)-(h) exist potential collisions. Hence, the robot needs to adjust their running parameters.  For scenario (e), the robot and the dynamic obstacle are running on the crossing paths and could meet at the crossing point, hence, the robot only needs to reduce its speed, thus allowing the dynamic obstacle to pass the potential collision point before the robot reaches the point, or increase its speed to enable itself to pass the potential collision point before the obstacle arrives at the point; for scenario (g), the robot can reduce its speed, make it keep a certain distance to the obstacle in front of the robot. The Morphin algorithm assumes that the robot is facing to the obstacle. Hence, it is applicable for scenarios (b) and (f). We can connect the robot's current position and the center of the obstacle to form a centerline, and draw several arcs on the left side and the right side of the centerline (Fig. \ref{fig:morphin}). Strictly,  the central line should be the link between the robot's position and the potential collision point (PCP) (red point in Fig. \ref{fig:scenarios}(a) and (e)).

As the robot is running in a grid environment, each of alternative arcs could occupy several grids. In the robot navigation, the path length is an important performance indicator. Moreover, the robot should not bear away from original path very much, hence, the turning angle should be as small as possible. Hence, the evaluation of each arc can be done in terms of the parameters of the arc in the grid environment, such as the arc length, the angle to the line from the robot to the target, as well as the distance between the path end and the sub-target. Wan et al. \cite{WanX2015} formulated the evaluation function of each arc as Eq. (\ref{eq:morphin}).
\begin{equation}\label{eq:morphin}
y =
\begin{cases}
\infty,  \text{if the path crosses an obstacle (a PCP)},\\
\varepsilon_1 L+\varepsilon_2 G+\varepsilon_3 \Delta L+\varepsilon_4 $W$, \text{others}.\\
\end{cases}
\end{equation}
where, $L$ is the length of each arc path, which is represented with the number of grids that the arc goes through from the start point (the place of robot) to the endpoint of the arc; $G$ is the parameter at the inflexion point on each arc path, and it is to ensure the robot does not go far away; $\Delta L$ is the average distance from each grid where the arc goes through to the sub-target on the global path, and it is to ensure the alternative path has a small distance to the sub-target; For scenarios (a) and (e), the sub-target could be set at the point of next grid close to the obstacle or collision point; $W = \frac{1}{1+m}$,  $m$ is the number of grids where both the arc and the global path go through; The parameter $W$ is to ensure the alternative path as close to the global path as possible. $\varepsilon_1$, $\varepsilon_2$, $\varepsilon_3$, $\varepsilon_4$ are the coefficients of the four items, respectively. If the arc crosses the obstacle, the value of $y$ is $\infty$, and the smallest $y$ indicates the alternative path is the best in the local probing area of the robot.\\

\begin{figure}
\centering
\includegraphics[height=2.2in]{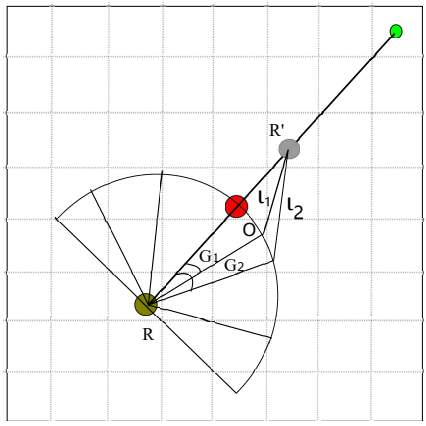}
\caption{Morphin paths on a grid map} \label{fig:morphingrid}
\end{figure}

For the simplicity of computing, the arcs can be drawn in straight lines (Fig. \ref{fig:morphingrid}), and the length of all paths beside the central line is the same as the length of the central line from the robot to the potential collision point.
Hence, the evaluation function could be simplified as Eq. (\ref{eq:sim-morphin}) regardless of the arc length $L$.
\begin{equation}\label{eq:sim-morphin}
y =
\begin{cases}
\infty,  \text{if the path crosses an obstacle (a PCP)},\\
\kappa_1 |G|+\kappa_2 \Delta L+\kappa_3 $W$, \text{others}.\\
\end{cases}
\end{equation}
where, $G$ is the angel between an alternative path and the central line (e.g. $G_1$, and $G_2$), $-\frac{\pi}{2}<G<\frac{\pi}{2}$; $\Delta L$ is the distance between the target and the sub-target ($R'$ in Fig. \ref{fig:morphingrid}), divided by the grid edge length; $W$ represents the number of times when the line between an alternative path endpoint and the sub-target point intersects the global path. Obviously, if the current position of the robot is on the global path (the straight line from start to the target), $W$ is at most 1/2.  If the robot is not on the global path, it is possible that a path does not intersect any grid on the global path, for which, $W$ =1. The larger the number of grids where the path intersects the global path, the smaller the value of $W$. $\kappa_1$...$\kappa_3$ are the weights of parameters in the evaluation function, respectively.
For example, as shown in Fig. \ref{fig:morphingrid}, $G_1<G_2$, $\Delta L_1 < \Delta L_2$, the number of grids where both $l_1$ and the global path go through is 4, while the number of grids where both $l_2$ and the global path go through is 3. Hence, $W_1=\frac{1}{4+1}=0.2$, $W_2= \frac{1}{3+1}=0.25$, and $W_1<W_2$. Obviously, $l_1$ is a better path than $l_2$, as $y_{l_1}$ < $y_{l_2}$.

There is no potential collision in scenarios (c) and (d). Hence, the robot will not change its moving parameters and continue. In scenario (e), the robot will stop until the obstacle passes the predicted collision point; in scenarios (b) and (f), a potential collision is identified, and the robot has to change its direction. Hence, the robot calls the Morphin algorithm to select the optimal alternative path in stead of the original path to avoid the potential collision; in scenario (g), the robot could slow down to avoid the potential collision. However, if the potential collision point has been in the sliding window, to avoid the rear-end accident, the robot either stops or call the Morphin algorithm to change the direction. Scenario (g) is different to Scenario (f), as the robot and the obstacle in Scenario (g) are moving in the same direction. In this research, for scenario (g), the robot will slow down or stop to avoid potential collision.

\section{Experiments and Evaluation}
The experiments include three parts: (1) the simulation experiments; (2) time performance assessment for different scenarios on a physical mobile robot, and (3) robustness test on a physical mobile robot in an environment with multiple obstacles.

\subsection{Simulation experiments}
The simulation is to verify the effectiveness of the proposed navigation system through some experiments. A specific start point and a specific target point are setup for robot's navigation tasks. All parameters in the experiments are set through the primary experiments with trial and error method. A $20\times20$ grid environment with many static obstacles is setup. The edge of each grid is set to 500mm. The radius $r_w$ of the sliding window is set to 8 grids.  As all obstacles added to the grid environment have a regular shape, the $\lambda$ value of adaptive threshold for data points clustering is set to 1.2; for simplicity, the values of thresholds in the decision tree (Fig.5) for obstacle recognition are set as: $\theta_{\varsigma_1}$=0.30, $\theta_{\varsigma_2}$=0.7, $\theta_\delta$=0.4, $\theta_\xi$=0.5, respectively. The parameters ($\kappa_1 \sim \kappa_3$) of Morphin Algorithm are set to  1, 1.3 and 0.6, the same as the corresponding values of $\varepsilon_2 \sim \varepsilon_4$ in \cite{WanX2015}.  Three experiments are conducted: (1) static obstacles only; (2) some dynamic obstacles appear in the grid environment; (3) static and dynamic obstacles appear instantly.

\subsubsection{Static obstacles in the environment}
\begin{figure}
\centering
\subfigure[Moving process 1:start point]
{\begin{minipage}{0.48\textwidth}\centering
\includegraphics[height=1.8in]{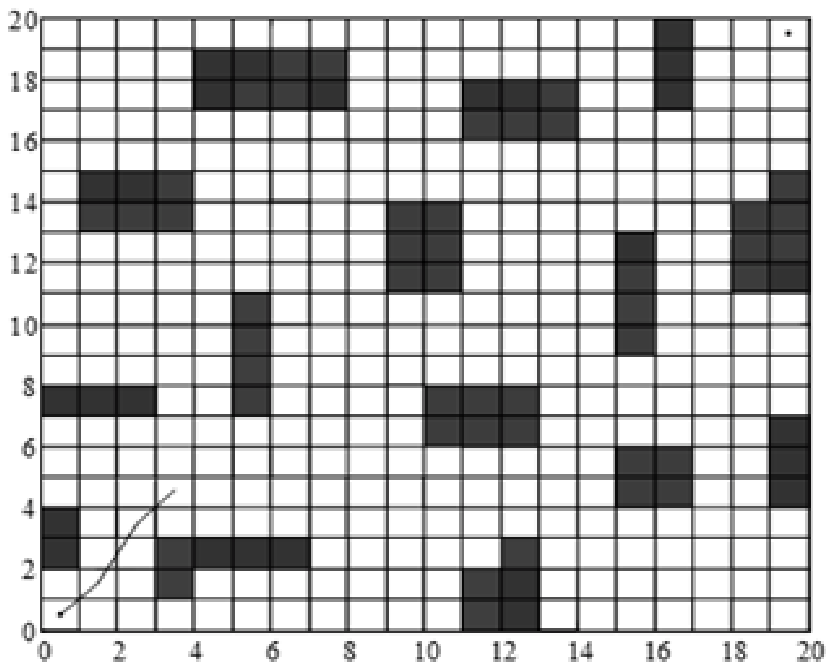}
\end{minipage}}
\subfigure[Moving process 2]
{\begin{minipage}{0.48\textwidth}\centering
\includegraphics[height=1.8in]{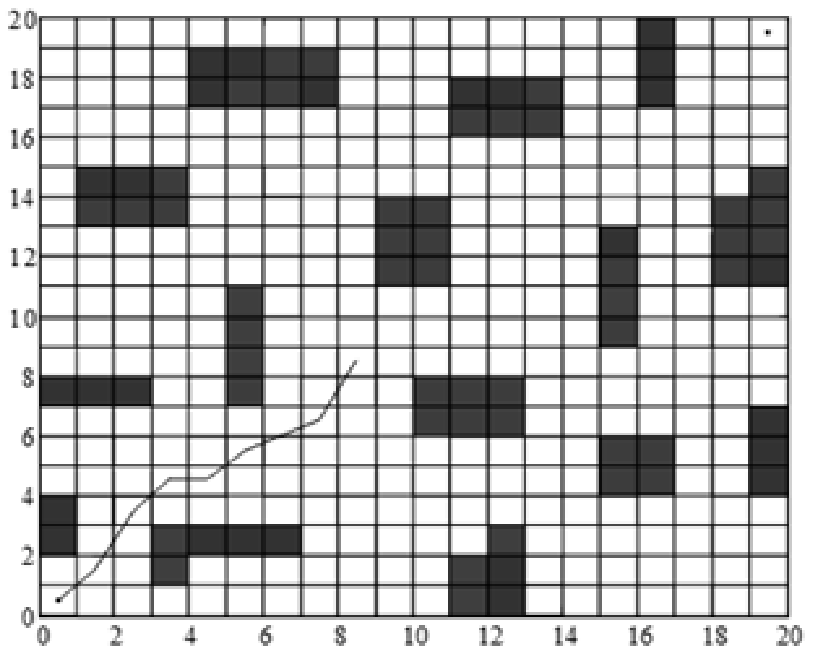}
\end{minipage}}
\subfigure[Moving process 3]
{\begin{minipage}{0.48\textwidth}\centering
\includegraphics[height=1.8in]{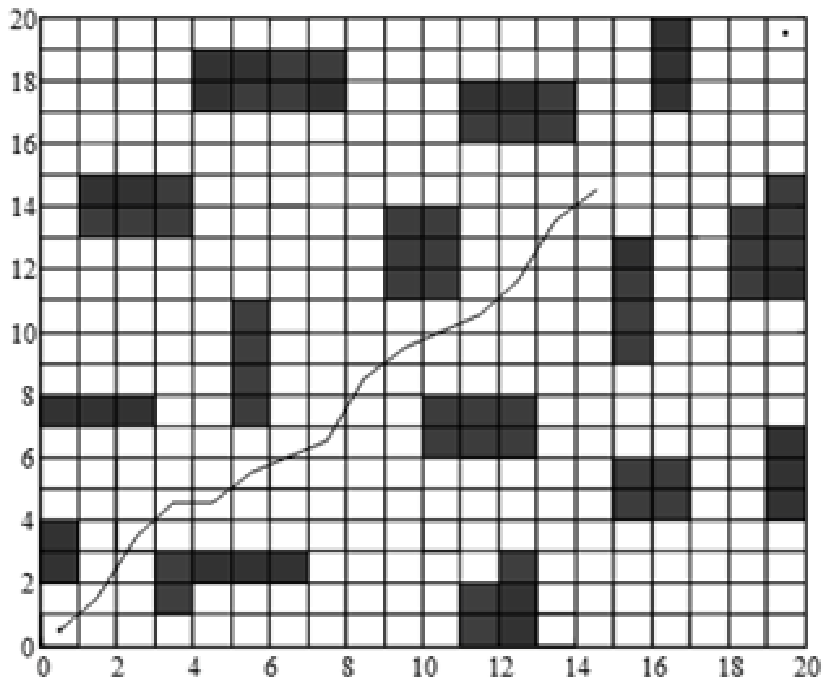}
\end{minipage}}
\subfigure[Moving process 4]
{\begin{minipage}{0.48\textwidth}\centering
\includegraphics[height=1.8in]{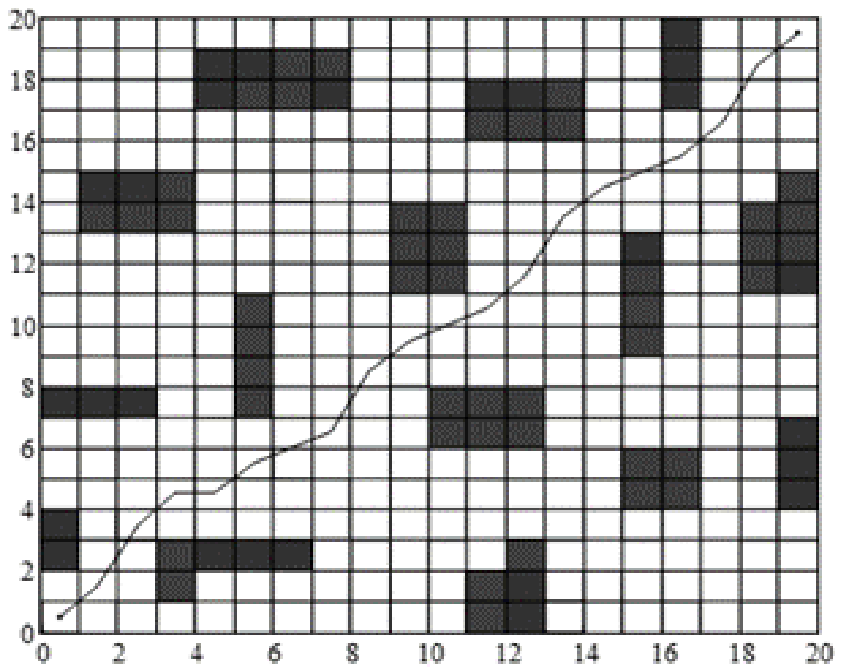}
\end{minipage}}
\caption{Local path planning in a static environment} \label{fig:staticenvironment}
\end{figure}

Assume there is a density of static obstacles in the experiment environment (Fig. \ref{fig:staticenvironment}) and no dynamic obstacles appear during the navigation of the robot. The start point is the bottom-left corner of the grid map, and the target is the top-right corner of the grid map. A mobile robot is moving from the specific start point to the specific target point in the unknown environment. The global optimal path is the straight line from the start to the target. During the navigation, if the mobile robot finds a static obstacle on its global path in the grid environment, it will call the Morphin path planning to avoid the obstacle towards next sub-target. In each step of the navigation, robot uses its current position as the centre to update the sliding window and moves towards the next step. The local sub-targets are updated towards the global target step by step, and constructs an actual trajectory from the start to the target (Fig. \ref{fig:staticenvironment}(a)-(d)). Finally, the robot completes the navigation. The experimental results demonstrate that the robot can effectively avoid the obstacles with multiple static obstacles on the global optimal path from start to the target.
\subsubsection{Instantly appeared dynamic obstacles in the environment}
\begin{figure}
\centering
\subfigure[Starting point: Adding ob1 ]
{\begin{minipage}{0.48\textwidth}\centering
\includegraphics[height=1.8in]{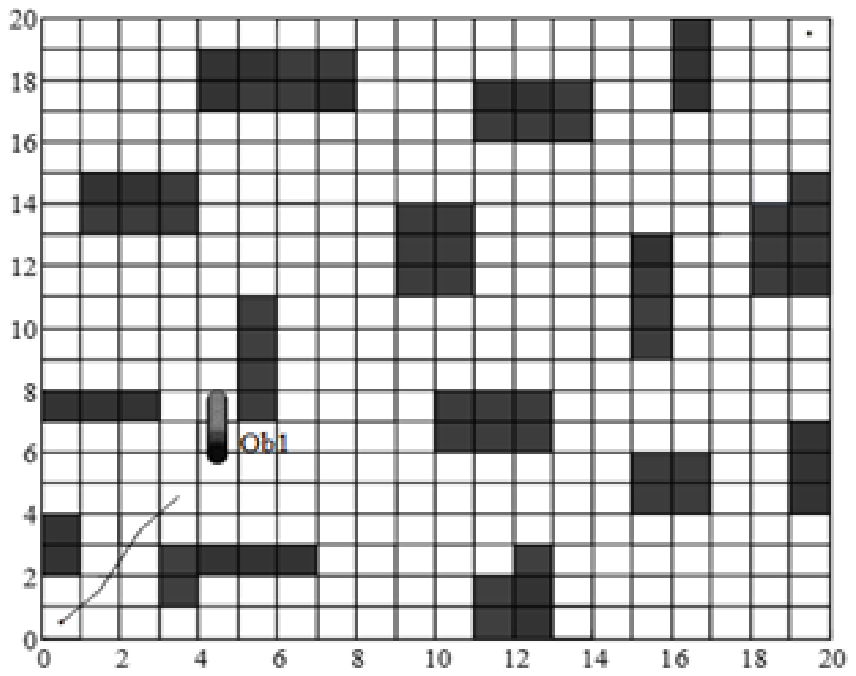}
\end{minipage}}
\subfigure[Avoiding Ob1, adding Ob2]
{\begin{minipage}{0.48\textwidth}\centering
\includegraphics[height=1.80in]{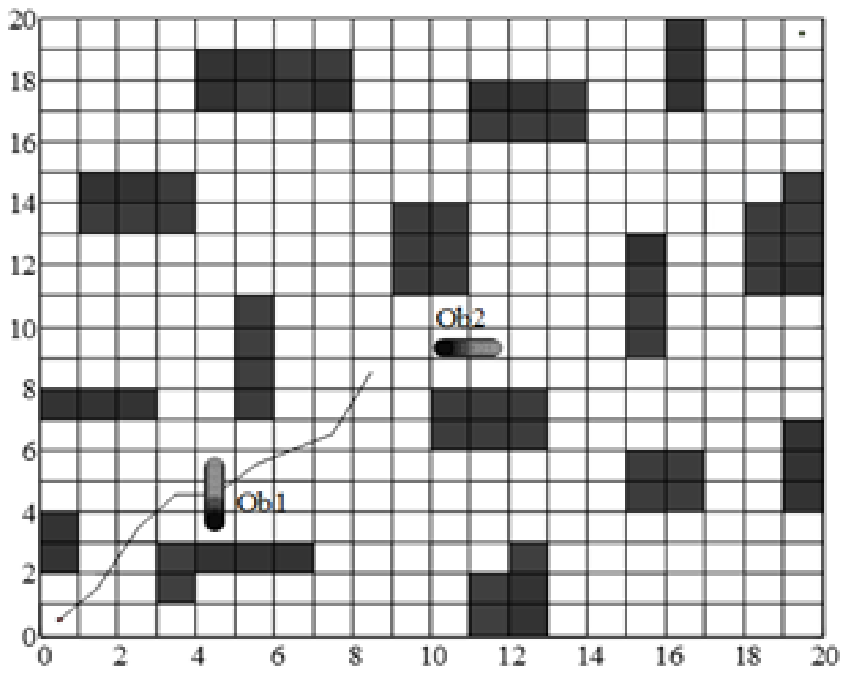}
\end{minipage}}
\newline
\subfigure[Avoiding Ob2, adding Ob3]
{\begin{minipage}{0.48\textwidth}\centering
\includegraphics[height=1.8in]{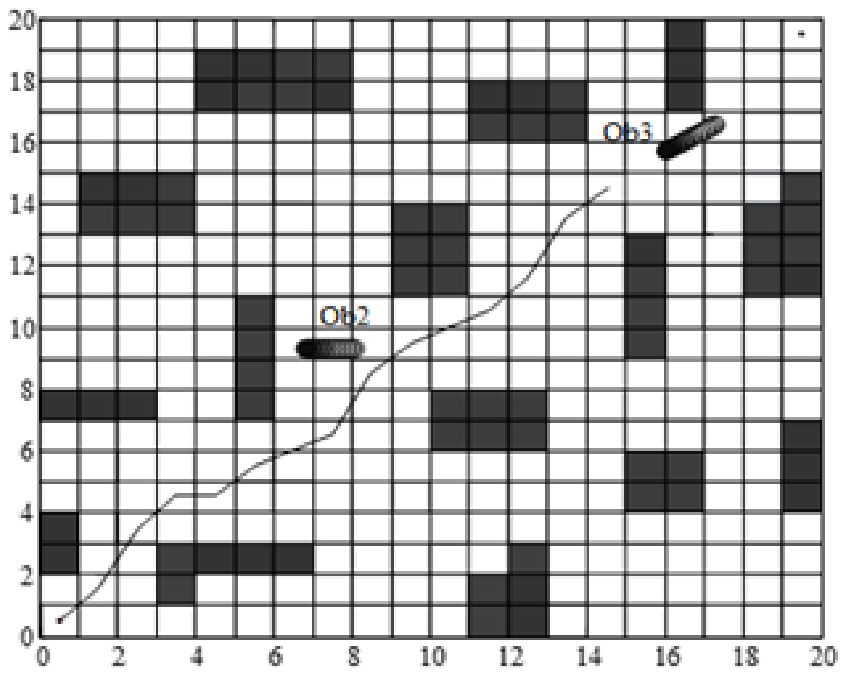}
\end{minipage}}
\subfigure[Avoiding Ob3, reaching the target]
{\begin{minipage}{0.48\textwidth}\centering
\includegraphics[height=1.8in]{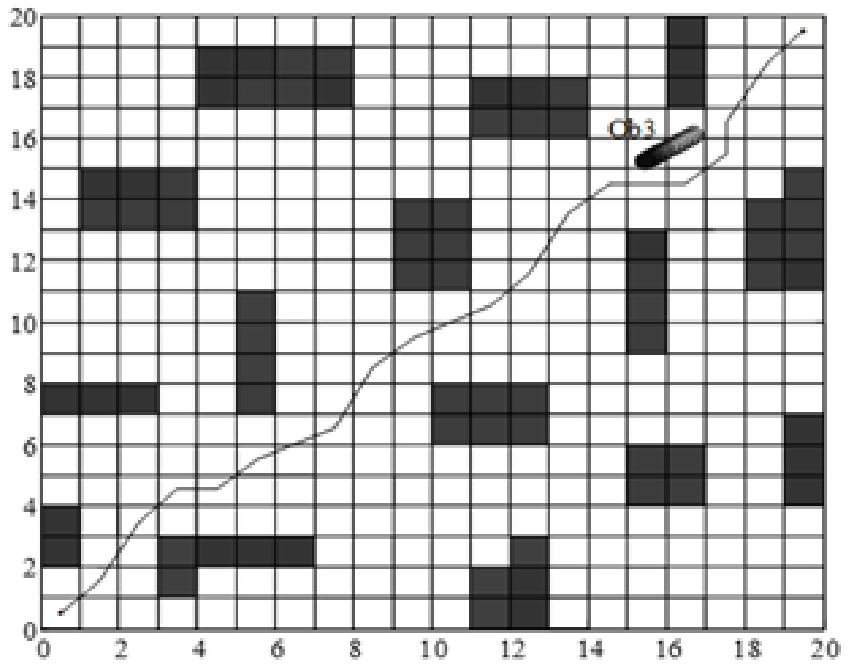}
\end{minipage}}
\caption{Local path planning in an environment with dynamic obstacles added instantly} \label{fig:dynamicenvironment}
\end{figure}

No matter what scenarios in which a robot is running, the robot will generate a local sub-target in each step. The computing complexity is increased for predicting potential collision point and path planning to avoid the dynamic obstacle.  Fig. \ref{fig:dynamicenvironment} shows the simulation of the local path planning process by adding three types of dynamic obstacles to the experimental environment. The mobile robot starts from the bottom-left corner, retrieves the data from the laser scanner, and identifies the dynamic obstacle $Ob_1$ in the current probing area. The robot calculates the motion parameters of $Ob_1$, and predicts that it will not reach the point where Ob1 crosses the path. This is Scenario (c) in Fig.\ref{fig:scenarios}. Hence, the robot will not collide with $Ob_1$, and it will keep the original speed and the moving direction. The mobile robot continues and finds $Ob_2$, a moving obstacle. The robot predicts that it will reach the point at the time when the obstacle crosses the path. Namely, the robot could collide with $Ob_2$, as shown in Scenario (e) of Fig.\ref{fig:scenarios}. Hence, it stops or slows down until $Ob_2$ passes the collision point (Fig. \ref{fig:dynamicenvironment} (c)). When a dynamic obstacle $Ob_3$ appears on the path that the robot is going through in the opposite direction towards the robot, as scenario (f) in Fig. \ref{fig:scenarios}, the collision is unavoidable if the robot does not change its direction. The robot predicts the potential collision point, calls the Morphin algorithm to get an optimal path, and moves to the sub-target, then follows the global path until reaching the final target, as shown in Fig. \ref{fig:dynamicenvironment} (d).
\subsubsection{A mixed case}
\begin{figure}
\centering
\includegraphics[height=3.6in]{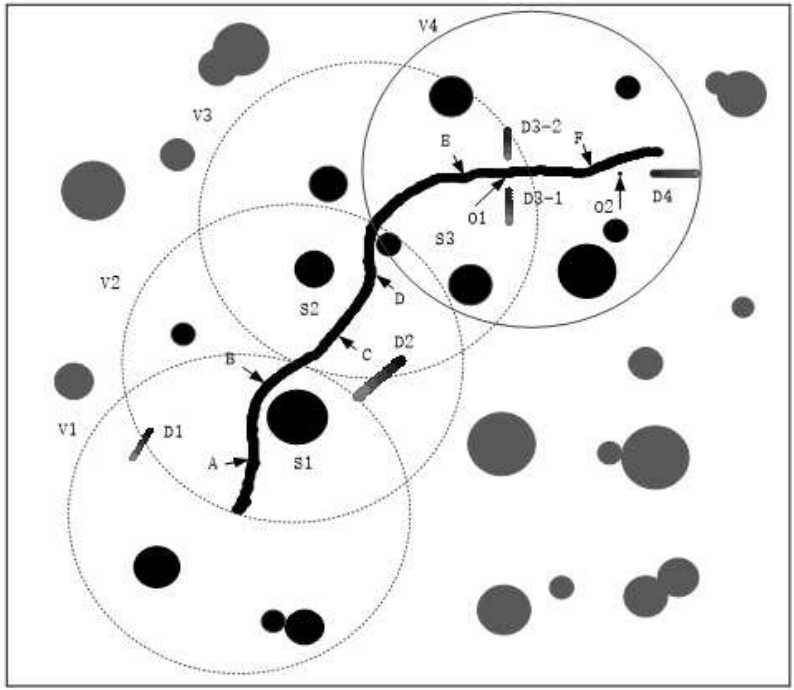}
\caption{Simulation results for the mix case} \label{fig:MixCase}
\end{figure}

Fig. \ref{fig:MixCase} illustrates the navigation process of instantly adding static and dynamic obstacles in the environment. In each step, the sliding window is updated in real time, and the trajectory of the robot actually forms the center trajectory of the sliding window. In Fig. \ref{fig:MixCase}, $V_1$, $V_2$, $V_3$ \& $V_4$ denote the sliding windows, $S_1$, $S_2$, $S_3$ \& $S_4$ are static obstacles in the environment, and $D_1$, $D_2$, $D_{3-1}$, $D_{3-2}$ \& $D_4$ are the dynamic obstacles appearing in the environment. The speeds and the angles of the four dynamic obstacles detected in the environment are shown in Table \ref{tab:paras}.
\begin{table}
\begin{center}
\caption{Movement speed and angle of obstacle}
\label{tab:paras}
\begin{tabular}{l|l|l|l|l}
\hline      &$D_1$          &$D_2$          &$D_3$          &$D_4$\\
\hline \hline
$v$	        &450mm/s	    &760mm/s	    &510mm/s	    &805mm/s\\
$\alpha$    &80.83$^\circ$  &62.47$^\circ$  &91.05$^\circ$  &180.09$^\circ$\\
\hline
\end{tabular}
\end{center}
\end{table}
As illustrated in Fig. \ref{fig:MixCase}, when the robot creates the sliding window $V_1$, at point $A$, it finds the static obstacle $S_1$, and executes the Morphin path planning to avoid the collision with $S_1$. The robot moves from $A$ to $B$, and finds the dynamic obstacle $D_1$ with the speed of 450mm/s and the angle of 80.83$^\circ$; It predicts that there is no potential collision with $S_1$, and continues toward the target; A sliding window of $V_2$ is created immediately after the robot runs beyond the probing area of $V_1$. When the robot moves from $C$ to $D$, it finds the static obstacle $S_2$ and the dynamic obstacle $D_2$ with the speed of 760mm/$s$ and the angle of 62.47$^\circ$, and predicts that there will be no collision with both of $S_2$ and $D_2$. At point $D$, the robot finds $S_3$ with potential collision, it executes the Morphin path planning again to avoid $S_3$; In the same way, when the robot is beyond the range of $V_2$, a sliding window of $V_3$ is updated. At point $E$, the robot finds the dynamic obstacle $D_{3-1}$ at a speed of 510 mm/s and the angle of 91.05$^\circ$, and predicts that it could collide with the obstacle at point $O_1$. The robot stops until the obstacle $D_{3-1}$ passes the potential collision point on the path to become $D_{3-2}$ in the environment. In the sliding window of $V_4$, at point $F$,  the robot finds the dynamic obstacle $D_4$ in the opposite direction towards the robot, at the speed of 805mm/s and the angle of 180.05$^\circ$. The robot predicts that the obstacle could collide with it at point $O_2$, if it does not change its path immediately. Hence, the robot executes the Morphin path planning immediately to select an optimal path, thus avoiding the collision with the obstacle $D_4$.
\subsection{Experiments for Time Performance Assessment}
\subsubsection{The Robot in the Experiment}
\begin{figure}
\centering
\subfigure[UP-VoyagerIIA]
{\begin{minipage}{0.48\textwidth}
\includegraphics[width=1.4in]{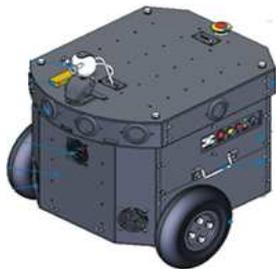}
\end{minipage}}
\subfigure[The lab environment]
{\begin{minipage}{0.48\textwidth}
\includegraphics[width=2.4in]{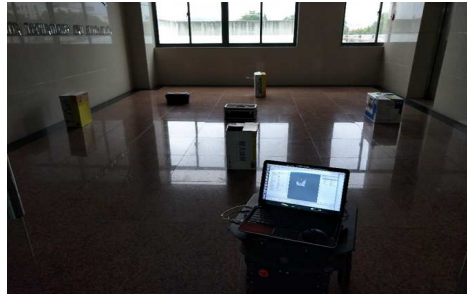}
\end{minipage}}
\caption{The UP-VoyagerIIA Robot and the Lab Environment} \label{fig:LabEnvironment}
\end{figure}

\begin{figure}
\centering
\includegraphics[height=3.2in]{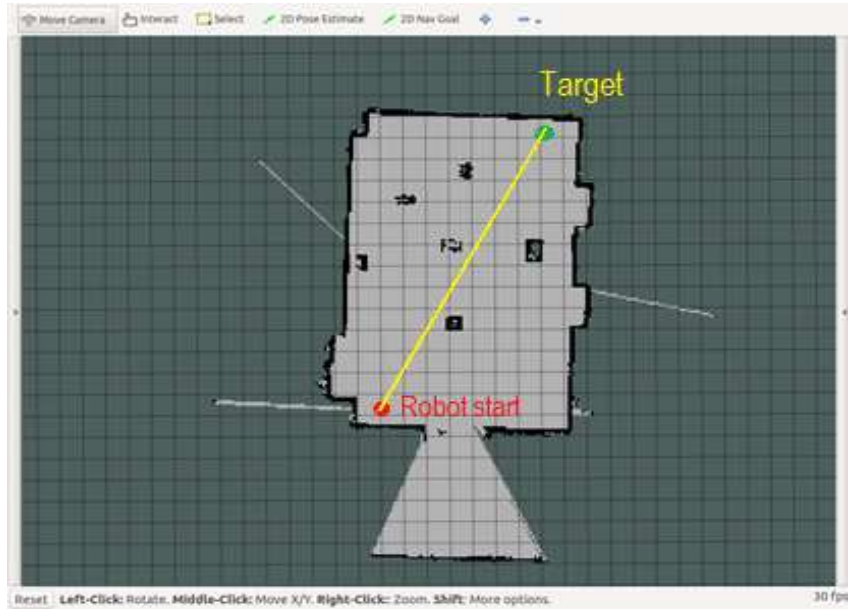}
\caption{The grid map of the robot lab environment} \label{fig:LabGripMap}
\end{figure}

 An ``UP-VoyagerII-A'' mobile robot (Fig. \ref{fig:LabEnvironment} (a)), the product of Beijing Bochuang Xingsheng Robot Technology Co., Ltd., is used for the performance test experiment. The robot is equipped with a SICK two-dimensional laser scanner, a twelve sonar ring, a Logitech camera and an odometer. But the data is collected from the laser sensor. The robot has an self-loading computer with i7-6700HQ, 4-Core processor, 16GB memory, GTX960M, and 2G discrete graphics. Table \ref{tab:specification} provides the specification of technical parameters of the robot.
\begin{table}
\begin{center}
\caption{Specification of Voyager-II}
\label{tab:specification}
\begin{tabular}{l|p{160pt}}
\hline
Item names            & Description\\
\hline \hline
SICK Laser scanner    & Cover 0~180$^\circ$, 0.5$^\circ$ resolution, 361 points in the range [0,180$^\circ$]\\
\hline
12 sonar sensors      & Cover 360$^\circ$, each sensor can detect obstacle within 30$^\circ$ cone\\
\hline
camera                & Logitech camera     \\
\hline
motor encoder         & 500 units per circle \\
\hline
Dell XPS15-9550-R4825 & control and communicate with the robot\\
\hline
\end{tabular}
\end{center}
\end{table}
On the Ubuntu 14.04 of the PC, ROS indigo is running. Fig. \ref{fig:LabEnvironment} (b) depicts the environment for the experiments of robot's navigation. The developed navigation system is embedded into the robotic system.
The robot is driven within the specified environment without any obstacle, and the environment information is collected. A high quality of grid map of the environment is produced and visualised in the window of the RVIZ system, as shown in Fig. \ref{fig:LabGripMap}.
\subsubsection{Experiments and Assessment}
The experiments are conducted for all the eight scenarios in Figs. \ref{fig:scenarios} (a) - (h) in the robotics lab. A mobile robot will start from the same point, automatically search and avoid obstacles on its path, and finally reach the same target. The initial speed of the robot $V_{R}=100mm/S$, and the straight-line distance between the start point and the target is 6$m$.  The start time is recorded when the mobile robot starts, and the end time is recorded when the robot reaches the target. In order to reduce the random errors, each experiment is repeated three times, and the average running time of the robot is calculated.  The experimental results are shown in Table \ref{tab:performance}.
\begin{table}
\centering
\caption{Approaching times of the robot for different scenarios}\label{tab:performance}
    \begin{tabular}{p{220pt}|l|l|l|l}
    \hline
    Scenarios & \multicolumn{3} {c|} {Approaching time(s)} & Av. (s) \\
    \hline\hline
    (a)	No obstacle                                 &61.77	&62.00	&61.98  &61.92\\
    \hline	
    (b) a static obstacle in front of the robot     &65.48	&64.37  &63.89  &64.58\\
    \hline	
    (c) a faster dynamic Obstacle                   &62.68	&62.27	&61.57	&62.17\\
    \hline	
    (d) a slower dynamic Obstacle                   &62.05	&62.24	&61.94	&62.07\\
    \hline	
    (e) Predicted collision with a dynamic Obstacle &64.53	&65.02	&64.97	&64.87\\
    \hline	
    (f) a dynamic obstacle in opposite direction towards the robot &76.12	&77.20	&76.45	&76.59\\
    \hline	
    (g) a slower dynamic obstacle in the same direction in front of the robot  &75.61	&76.01	&75.92	&75.85\\
    \hline	
    (h) mix obstacles	&77.85	&78.56	&78.97	&78.46\\
    \hline
    \end{tabular}
\end{table}
Experimental results show: for scenario (a), the robot does not meet any obstacle, thus directly reaches the target with smallest time; for scenario (b), the robot finds a static obstacle, successfully avoids the obstacle, and reaches the target; for scenario (c), the robot finds a dynamic obstacle, which has a faster speed than the robot, the robot predicts the obstacle will pass the path before it reaches the potential collision point, keeps the original speed and direction, and safely reaches the target; for scenario (d), the robot finds a slower obstacle, the robot predicts that it will not conflict with the detected obstacle if the robot keeps the original speed and the direction; for scenarios (e), (f) and (g), the robot must adjust moving direction or speed, it will conflict with the obstacle. For scenario (e), the robot either speed up or slow down to avoid the obstacle; for scenario (f), the robot must change direction, so the robot applies the Morphin algorithm and successfully avoids the obstacle; for scenario (g), the robot reduces its speed, so that it is slower than the obstacle in its front; The experimental results demonstrate that the robot can continuously deal with different obstacles in its front.

From the results in Table \ref{tab:performance}, it can be concluded that:
\begin{itemize}
\item [(1)] the results for scenario (a) shows the robot spent the shortest time to reach the target, namely, for all other scenarios, the robot needs to spend some time to detect and avoid obstacles, and select a proper path.
\item [(2)] for scenarios (c)and (d),although the robot does not need to change its moving direction and speed to reach the target, it still needs to spend time to detect the obstacle and make the decision. For these two scenarios, the running times of the robot are similar, but they are shorter than that for scenarios (b) and (e), in which, the robot needs to run the Morphin algorithm to select a suitable path.
\item [(3)] for scenarios (b) and (e), the running time of the robot is for obstacle detection, obstacle classification, collision prediction and path planning. Obviously the running time is longer than that in Scenarios (c) (d). The average time for static obstacle (Scenario (b)) is slightly less than that for dynamic obstacle (Scenario (e)).
\item [(4)] for scenario (f), as the obstacle is running in the opposite direction towards the robot, the robot detects the obstacle multiple times, and even after the robot changes its direction, the obstacle is still in the range of the sliding window. Hence, the running time is longer than Scenarios (a) - (e).
\item [(5)] for scenario (g), the obstacle is running in the same direction as the robot, the robot is faster than the obstacle and follows after the obstacle, hence, the distance between the robot and the obstacle is getting closer and closer. In this scenario, the robot detects the obstacle many times, and once the robot identifies that it could collide with the obstacle, it will slow down and keep a certain distance to the obstacle. Hence, the running time of the obstacle is longer than scenarios (a) - (e), but slightly shorter than scenario (f), as in Scenario (g), the robot does not call the Morphin function for path planning.
\item[(6)] In scenario (h), the robot needs to deal with two obstacles: a static obstacle and a dynamic obstacle. Hence, the robot continues detecting obstacles on the path towards the target. When the robot finds out the dynamic obstacle, it reduces its speed to avoid the potential concision; when the robot finds out the static obstacle on the path, the robot needs to call the Morphin function for path planning. Therefore, the running time of the robot, reaching the target, is longest in all scenarios.
\end{itemize}
\subsection{Robustness of the Navigation System in an Environment with Multiple Obstacles}
The robot, used for the validation of robustness, is the same as in Experiment 2, and the experiments are conducted in the same lab environment,but with multiple static or dynamic obstacles. Robot's speed $V_R$ = 100mm/s, $V_{Ob1}$=0 mm/s, $V_{Ob2}$=75 mm/s, $V_{Ob3}$=150 mm/s, $V_{Ob4}$=100 mm/s.

First, in the way as in Experiment 2, a grid map of the lab environment is created.  Secondly, the start and the end points of robot navigation are setup. Under the static environment without dynamic obstacles, the robot can run along with the optimal path, the straight line from the start point to the target, shown as the yellow line in the grid map Fig. \ref{fig:LabGripMap}.

\begin{figure*}
\subfigure[Starting point: Static ob1 detected]
{\begin{minipage}{0.48\textwidth}\centering
\includegraphics[height=1.8in]{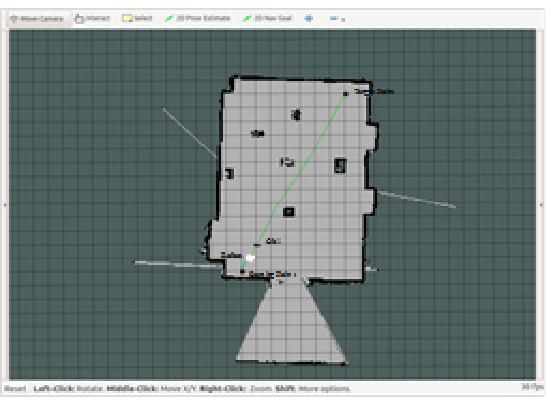}
\end{minipage}}
\subfigure[Avoiding static Ob1, dynamic Ob2 detected]
{\begin{minipage}{0.48\textwidth}\centering
\includegraphics[height=1.8in]{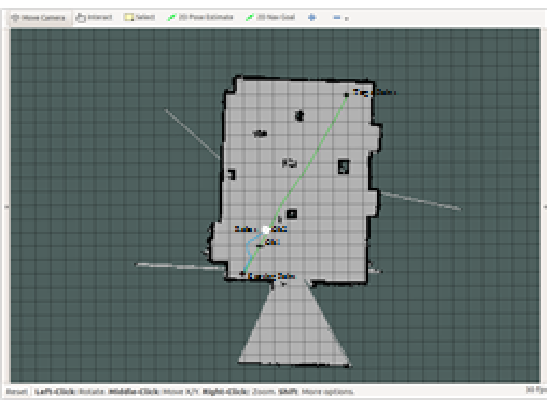}
\end{minipage}}
\subfigure[keep moving with original speed, avoiding Ob2]
{\begin{minipage}{0.48\textwidth}\centering
\includegraphics[height=1.8in]{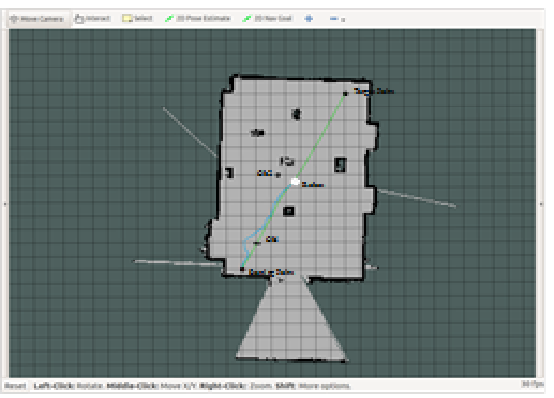}
\end{minipage}}
\subfigure[Dynamic Ob3 detected]
{\begin{minipage}{0.48\textwidth}\centering
\includegraphics[height=1.8in]{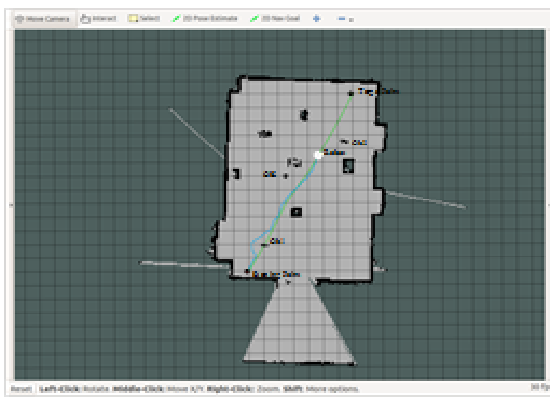}
\end{minipage}}
\subfigure[Avoiding Ob3, dynamic Ob4 detected]
{\begin{minipage}{0.48\textwidth}\centering
\includegraphics[height=1.8in]{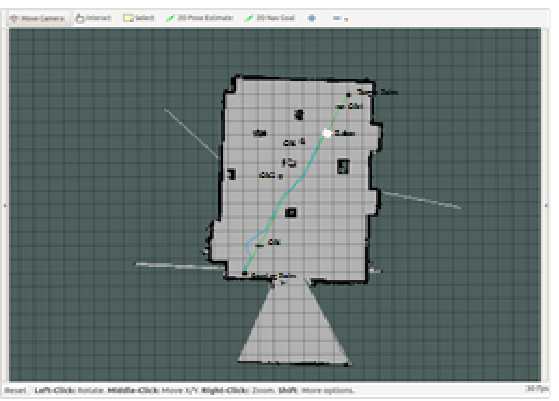}
\end{minipage}}
\subfigure[Avoiding Ob4, reaching the target]
{\begin{minipage}{0.48\textwidth}\centering
\includegraphics[height=1.8in]{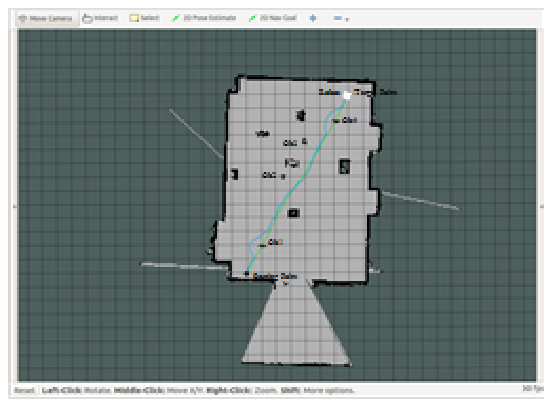}
\end{minipage}}
\caption{Local path planning in the lab environment} \label{fig:LabExps}
\end{figure*}

In the lab environment, adding some temporary static and dynamic obstacles, the mobile robot detects obstacles through the laser scanner, classifies obstacles, predicts the state of obstacles, and avoids them on the path. Fig. \ref{fig:LabExps} depicts the process of robot navigation from the start to the target. The green line is the optimal line, and the blue line is the real  path that the robot executes.
As shown in Fig. \ref{fig:LabExps} (a), the actual moving path of the mobile robot is along with the blue line, while the static and dynamic obstacles are added successively. The mobile robot moves along the global path from the start point. When finding a static obstacle $Ob_1$ appeared on the global path (Scenario (b)), the robot calls the Morphin algorithm, and changes its path.

After safely avoiding the obstacle $Ob_1$, the robot gets back to the global path; after moving for a distance, a dynamic obstacle $Ob_2$ is detected (Fig. \ref{fig:LabExps} (b)), and the potential collision is predicted to occur at the front of the robot (Scenario (e)). The obstacle avoidance strategy is that the mobile robot stops in the place until the obstacle passes the path and leaves away  (Fig. \ref{fig:LabExps} (c)).

As shown in Fig. \ref{fig:LabExps} (d), when the dynamic obstacle $Ob_3$ is detected, the mobile robot predicts the potential collision and judges that the original speed can be maintained (Scenario (d));

As show in Fig. \ref{fig:LabExps} (e), when the obstacle $Ob_4$ is detected, the predicted collision is inevitable (Scenario (f)), hence, the robot carried out the Morphin algorithm immediately, and changes its path before the collision to avoid the obstacle, and then goes back the global path, and finally, it reaches the target (Fig. \ref{fig:LabExps} (f)).

Briefly, the experimental results have proved that the proposed navigation system is effective. As the computing is simple, the navigation can perform in real-time. It can be seen that, for Scenarios (b) and (f), the robot can effectively avoid the potential collision if it changes its path, using the Morphin path planning; for Scenarios (e) and (g), the robot can avoid the potential collision by stopping or slowing down its speed; for Scenario (g), if the robot needs to overtake the slowly moving obstacle in its front, then it needs to call the Morphin algorithm to change its path, which is similar to the Scenario (f). But it should be noticed that the sub-target, at which, the robot goes back to the global path for Scenario (g) should be further than the sub-target on the global path for Scenarios (b) and (f), as the obstacle in Scenario (g) is running on the same direction as the robot.
\section{Conclusions}
We have developed a new navigation algorithm, which borrows the concept of polar histograms in VFH for obstacle detection, using an adaptive threshold clustering algorithm, classifies the detected obstacles, predicts the potential collision and finds the optimal path with the simplified Mophin algorithm. The computing complexity of the proposed algorithm is good, as it avoids the training process of machine learning model, thus implementing the full autonomy of a robot with a fast and effective navigation in an unknown environment. The eight scenarios are analysed and tested in simulation and on a physical robot. The experimental results demonstrate that the proposed navigation system enables a mobile robot to effectively and efficiently avoid any static and dynamic obstacles on the path, where the robot goes through. As the navigation system was implemented with a good modularity, it is easy to replace any modules in the system with other algorithms. This could provide a good platform for future research.

The study of scenario (g), a robot is faster than a dynamic obstacle ahead of the robot in the same direction, could be applied for autonomous vehicles to avoid the potential rear-end accidents, which is a critical challenge in autonomous vehicles. Further research on scenario (g), overtaking a dynamic obstacle to avoid the rear-end accident, the verification of the maximal speed of the robot in different cases in a real environment, as well as the improvement of the real-time performance of the navigation system, will be the future work.

\end{document}